# A Global Benchmark of Algorithms for Segmenting Late Gadolinium-Enhanced Cardiac Magnetic Resonance Imaging

Zhaohan Xiong[1], Qing Xia[2], Zhiqiang Hu[3], Ning Huang[4], Cheng Bian[5], Yefeng Zheng[5], Sulaiman Vesal[6], Nishant Ravikumar[6], Andreas Maier[6], Xin Yang[7], Pheng-Ann Heng[7], Dong Ni[8], Caizi Li[9], Qianqian Tong[9], Weixin Si[10], Elodie Puybareau[11], Younes Khoudli[11], Thierry Géraud[11], Chen Chen[12], Wenjia Bai[12], Daniel Rueckert[12], Lingchao Xu[13], Xiahai Zhuang[14], Xinzhe Luo[14], Shuman Jia[15], Maxime Sermesant[15], Yashu Liu[16], Kuanquan Wang[16], Davide Borra[17], Alessandro Masci[17], Cristiana Corsi[17], Coen de Vente[18], Mitko Veta[18], Rashed Karim[19], Chandrakanth Jayachandran Preetha[20], Sandy Engelhardt[21], Menyun Qiao[22], Yuanyuan Wang[22], Qian Tao[23], Marta Nuñez-Garcia[24], Oscar Camara[24], Nicolo Savioli[25], Pablo Lamata[25], Jichao Zhao[1]

[1]*Auckland Bioengineering Institute, University of Auckland, Auckland, New Zealand.*
[2]*State Key Lab of Virtual Reality Technology and Systems, Beihang University, Beijing, China.*
[3]*School of Electronics Engineering and Computer Science, Peking University, Beijing, China.*
[4]*SenseTime Inc, Shenzhen, China.*
[5]*Tencent Jarvis Laboratory, Shenzhen, China.*
[6]*Pattern Recognition Lab, Friedrich-Alexander-Universität Erlangen-Nürnberg, Erlangen, Germany.*
[7]*Department of Computer Science and Engineering, The Chinese University of Hong Kong, Hong Kong.*
[8]*National-Regional Key Technology Engineering Laboratory for Medical Ultrasound, School of Biomedical Engineering, Health Science Center, Shenzhen University, Shenzhen, China.*
[9]*School of Computer Science, Wuhan University, Wuhan, China.*
[10]*Shenzhen Key Laboratory of Virtual Reality and Human Interaction Technology, Shenzhen Institutes of Advanced Technology, Chinese Academy of Sciences, Shenzhen, China.*
[11]*EPITA Research and Development Laboratory, Paris, France.*
[12]*Department of Computing, Imperial College London, London, United Kingdom.*
[13]*School of Naval Architecture, Ocean & Civil Engineering, Shanghai Jiao Tong University, Shanghai, China.*
[14]*School of Data Science, Fudan University, Shanghai, China.*
[15]*Universite Cote d'Azur, Epione Research Group, Inria, Sophia Antipolis, France.*
[16]*School of Computer Science and Technology, Harbin Institute of Technology, Harbin, China.*
[17]*Department of Electric, Electronic and Information Engineering, University of Bologna, Cesena, Italy.*
[18]*Department of Biomedical Engineering, Eindhoven University of Technology, Eindhoven, The Netherlands.*
[19]*School of Biomedical Engineering & Imaging Sciences, Kings College London, London, United Kingdom.*
[20]*Faculty of Electrical Engineering and Information Technology, University of Magdeburg, Magdeburg, Germany.*
[21]*Heidelberg University Hospital, Germany.*
[22]*Biomedical Engineering Center, Fudan University, Shanghai, China.*
[23]*Department of Radiology, Leiden University Medical Center, Leiden, the Netherlands.*
[24]*Physense, Department of Information and Communication Technologies, Universitat Pompeu Fabra, Barcelona, Spain.*
[25]*Department of Bioengineering, Kings College London, London, United Kingdom.*

Corresponding Author: Dr. Jichao Zhao

Address: Auckland Bioengineering Institute, The University of Auckland, Private Bag 92019, Auckland 1142, New Zealand.
Email: j.zhao@auckland.ac.nz.

**Abstract**

Segmentation of cardiac images, particularly late gadolinium-enhanced magnetic resonance imaging (LGE-MRI) widely used for visualizing diseased cardiac structures, is a crucial first step for clinical diagnosis and treatment. However, direct segmentation of LGE-MRIs is challenging due to its attenuated contrast. Since most clinical studies have relied on manual and labor-intensive approaches, automatic methods are of high interest, particularly optimized machine learning approaches. To address this, we organized the "2018 Left Atrium Segmentation Challenge" using 154 3D LGE-MRIs, currently the world's largest cardiac LGE-MRI dataset, and associated labels of the left atrium segmented by three medical experts, ultimately attracting the participation of 27 international teams. In this paper, extensive analysis of the submitted algorithms using technical and biological metrics was performed by undergoing subgroup analysis and conducting hyper-parameter analysis, offering an overall picture of the major design choices of convolutional neural networks (CNNs) and practical considerations for achieving state-of-the-art left atrium segmentation. Results show the top method achieved a dice score of 93.2% and a mean surface to a surface distance of 0.7 mm, significantly outperforming prior state-of-the-art. Particularly, our analysis demonstrated that double, sequentially used CNNs, in which a first CNN is used for automatic region-of-interest localization and a subsequent CNN is used for refined regional segmentation, achieved far superior results than traditional methods and pipelines containing single CNNs. The findings from this study can potentially be extended to other imaging datasets and modalities, having an impact on the wider medical imaging community. This large-scale benchmarking study makes a significant step towards much-improved segmentation methods for cardiac LGE-MRIs, and will serve as an important benchmark for evaluating and comparing the future works in the field.

# 1. Introduction

*1.1. Background*

Segmentation is an important task for the quantitative analysis of medical images. In particular, delineation of a patient's internal organ and tissue structure from 3D images, such as those obtained from computed tomography (CT) and magnetic resonance imaging (MRI), is often a necessity for medical diagnosis, patient stratification, and clinical treatment (Medrano-Gracia et al. 2015, Oakes et al. 2009, Csepe et al. 2017). Nowadays, various contrast agents are widely used to improve the visibility of disease-associated structures such as scarring, tumors, and blood vessels. For example, gadolinium-based contrast-enhancing agents are used in a third of all MRI scans (LGE-MRIs) worldwide and are proved to be very effective in providing clinical diagnosis of cardiac diseases (Oakes et al. 2009, Higuchi et al. 2017, Hennig et al. 2017, Figueras i Ventura et al. 2018). However, direct segmentation and analysis of cardiac LGE-MRIs remain challenging due to the attenuated contrast in nondiseased tissue and imaging artifacts, as well as the varying quality of imaging. Therefore, the current standard of image segmentation and 3D reconstruction from these images for medical use still relies heavily on labor-intensive manual or semi-automatic methods (Csepe et al. 2017, Oakes et al. 2009, Higuchi et al. 2017). For instance, researchers and clinicians at the University of Utah have utilized LGE-MRIs for cardiac research in the past two decades based on their well-established, although time-consuming workflow to manually label the left atrium (LA) from LGE-MRIs in patients with atrial fibrillation, the most common cardiac arrhythmia (McGann et al. 2014).

*1.2. Related Work*

In recent years, many approaches have been proposed for performing automatic 3D segmentation of cardiac structures from medical images, mostly for non-contrast data. A 2013 benchmarking study held in conjunction with Medical Image Computing and Computer Assisted Intervention (*MICCAI*) examined methods for automatically segmenting the LA from non-contrast MRIs and CTs (N = 30 each) (Tobon-Gomez et al. 2015). The study revealed that methods combining statistical or atlas-based models with region growing approaches performed the best for both image types among 9 participants. A more recent 2017 benchmarking study for segmentation of left and right ventricles from non-enhanced MRIs showed that convolutional neural networks (CNNs) significantly outperformed traditional methods (Bernard et al. 2018). By analyzing algorithms from 10 participants, the study revealed that the popular U-Net CNN architecture (Ronneberger, Fischer and Brox 2015) which was specifically designed for medical image segmentation was particularly effective. Contrary to non-enhanced images, contrast-enhanced MRIs/CTs have received significantly less attention for research despite their important role in clinics. A 2016 study was conducted to investigate methods of LA segmentation from LGE-MRIs and contrast-enhanced CTs (N = 10 each) by analyzing the submitted algorithms from three groups (Karim et al. 2018). However, due to the limited number of submitted algorithms, the study was not able to draw any definitive conclusions in terms of approach development. Other studies on LGE-MRI segmentation also have limited efficacy as the methods proposed require additional information such as manually initialized shape priors (Veni, Elhabian and Whitaker 2017, Zhu et al. 2013) or paired magnetic resonance angiography (Tao et al. 2016). While LGE-MRI segmentation still heavily relies on traditional methods, many recent advancements in CNNs have been made for image segmentation in general. VGGNet (Simonyan and Zisserman 2014) has been widely used for developing full convolutional networks (Long, Shelhamer and Darrell 2015, Noh, Hong and Han 2015) for semantic segmentation due to its simplicity, and adaptations of superior architectures, such as ResNet and Inception (Szegedy et al. 2017), are currently the state-of-the-art in the field (Chen et al. 2018b). Thus, the application of current cutting-edge CNN architectures to LGE-MRIs requires further investigation and development.

*1.3. Contribution*

It is unknown if it is possible to create a unified deep learning architecture capable of achieving optimal performance for segmentations across a wide spectrum of applications. While U-Net, VGGNet, and ResNet are currently the most widely used CNN architectures for medical image segmentation, CNNs still have to be individually tuned for each specific application (Knoll, Maier and Rueckert 2018, Zhu 2019). This is reflected in the current literature which often contains wildly differing implementations of U-Net and other architectures for segmentation tasks in different disciplines, making it difficult to pinpoint design characteristics that can be applied universally. Methods for hyper-parameter optimization, particularly in U-Net, for general problems and/or specific tasks are still an ongoing topic of discussion, and could potentially lead to a more robust framework for segmentation. The lack of accessible, large-scale datasets and the varying quality of the image data (with labels) also hinders the development of optimized methods for LA segmentation from LGE-MRIs. By providing the largest 3D cardiac LGE-MRI dataset along with top quality expert labeled LA cavities ($N = 154$) thanks to the great efforts of University of Utah over the past two decades, we have gathered the community and organized the 2018 LA Segmentation Challenge (**Figure 1**) in conjunction with the MICCAI conference and Statistical Atlases and Computational Modelling of the Heart (STACOM) workshop in Granada, Spain (Pop 2019). Throughout the course of the challenge, over 200 research groups/individuals worldwide accessed our challenge dataset. In total, 27 teams participated in the final evaluation phase of the challenge and their final rankings are shown on the challenge website (http://atriaseg2018.cardiacatlas.org/).

To analyze the wide spectrum of conventional and deep learning methods submitted to the challenge, we first sub-grouped the methods by their main architecture designs and then identified the subgroup with the best design features which contributed to its superior performance. We then conducted extensive hyper-parameter tuning experiments on the top-performing method to identify the exact parameter choices leading to achieving state-of-the-art accuracies. By performing the benchmarking study in this manner, we offered both an overall picture on the major design choices necessary, as well as detailed practical considerations.

Through our global benchmarking study, we aim to derive an optimized framework for the segmentation of the LA in LGE-MRIs, providing critical insights into effective approaches and techniques for segmentation. These approaches can potentially be extended to other areas of cardiac segmentation, as well as for non-contrast MRIs, potentially having a high impact on the wider imaging community.

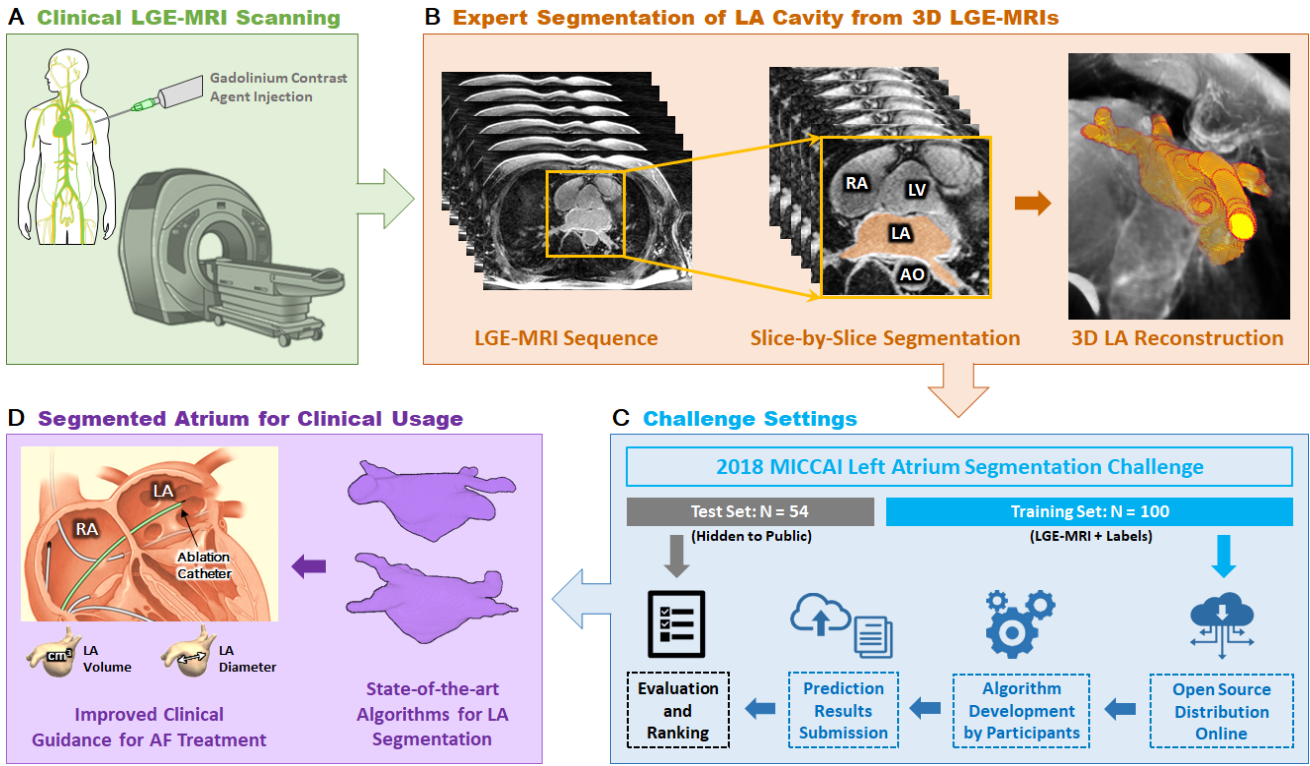

**Figure 1:** Overall workflow of medical images for clinical usage and the 2018 Left Atrium (LA) Segmentation Challenge. **A)** Clinical MRI scanners were used to acquire late gadolinium-enhanced magnetic resonance imaging (LGE-MRIs) from patients with atrial fibrillation (AF). **B)** The LGE-MRIs were manually segmented in a slice-by-slice manner by experts to obtain labels of the LA cavity. The 3D LA geometry can be obtained by stacking the 2D segmentation together. **C)** In the 2018 LA Segmentation Challenge, 154 3D LGE-MRI data (each with a spatial dimension of either 576×576×88 or 640×640×88) was split into 100 training and 54 testing sets. The training data and labels were made public to all potential participants of the challenge, and the testing data was used at the end of the challenge for evaluation. A total of 27 teams participated and were ranked based on the Dice scores. **D)** Accurate reconstruction of the LA anatomical structure provides crucial information for patient stratification and for guiding clinical treatment for patients with AF. AO, aorta; LV, left ventricle; RA, right atrium.

## 2. Methods and Materials
### 2.1. Dataset and Labels

A total of 154 independently acquired 3D LGE-MRIs from 60 de-identified patients with atrial fibrillation were used in this challenge. All patients underwent LGE-MRI scanning to define the atrial structure and fibrosis distribution prior to and post-ablation treatment at Utah (McGann et al. 2014, McGann et al. 2011). High-resolution LGE-MRIs of bi-atrial chambers were acquired approximately 20-25 minutes after the injection of 0.1 mmol/kg gadolinium contrast (Multihance, Bracco Diagnostics Inc., Princeton, NJ) using a 3D respiratory navigated, inversion recovery prepared gradient echo pulse sequence. An inversion pulse was applied every heartbeat, and fat saturation was applied immediately before data acquisition. To preserve magnetization in the image volume, the navigator was acquired immediately after the data acquisition block. Typical scan times for the LGE-MRI study were between 8-15 minutes at 1.5 T and 6-11 minutes using the 3T scanner (for Siemens sequences) depending on patient respiration (McGann et al. 2014, McGann et al. 2011).

The spatial resolution of one 3D LGE-MRI scan was 0.625×0.625×0.625 mm³ with spatial dimensions of either $576 \times 576 \times 88$ or $640 \times 640 \times 88$ pixels. The LA cavity volumes were manually segmented in consensus with three trained observers for each LGE-MRI scan using the Corview image processing software (Merrk Inc, Salt Lake City, UT) (McGann et al. 2014). The LA cavity was defined as the pixels contained within the LA endocardial surface including the mitral valve and LA appendage as well as an extent of the pulmonary vein (PV) sleeves. The endocardial surface border of the LA was segmented by manually tracing the LA and PV blood pool which were regions with enhanced pixel intensities in each slice of the LGE-MRI volume. The extent of the PV sleeves in the endocardial segmentations was limited to the PV antrum region, and was defined as the point where the PVs stopped narrowing and remained constant in diameter. On average, the PV antra were limited to less than 10 mm extending out from the endocardial surface, or approximately three times the thickness of the LA wall. The LGE-MRI image volumes and associated LA segmentations were stored in the nearly-raw raster data (nrrd) format. The LGE-MRIs were grayscale and the segmentation labels were binary.

In order to validate the quality of the dataset, the signal to noise ratio (SNR), contrast ratio (CR), and heterogeneity (HET) between the foreground containing the LA and the background was assessed (**Figure 2A-C**). The three metrics used were also in agreement as SNR had a strong positive correlation with CR and HET and while CR and HET had a strong negative correlation. Distributions of the quality measurements on all data showed that less than 15% of the data was of high quality (SNR < 1), 70% of the data was of medium quality (SNR = 1 to 3), and over 15% of the data was of low quality (SNR > 3).

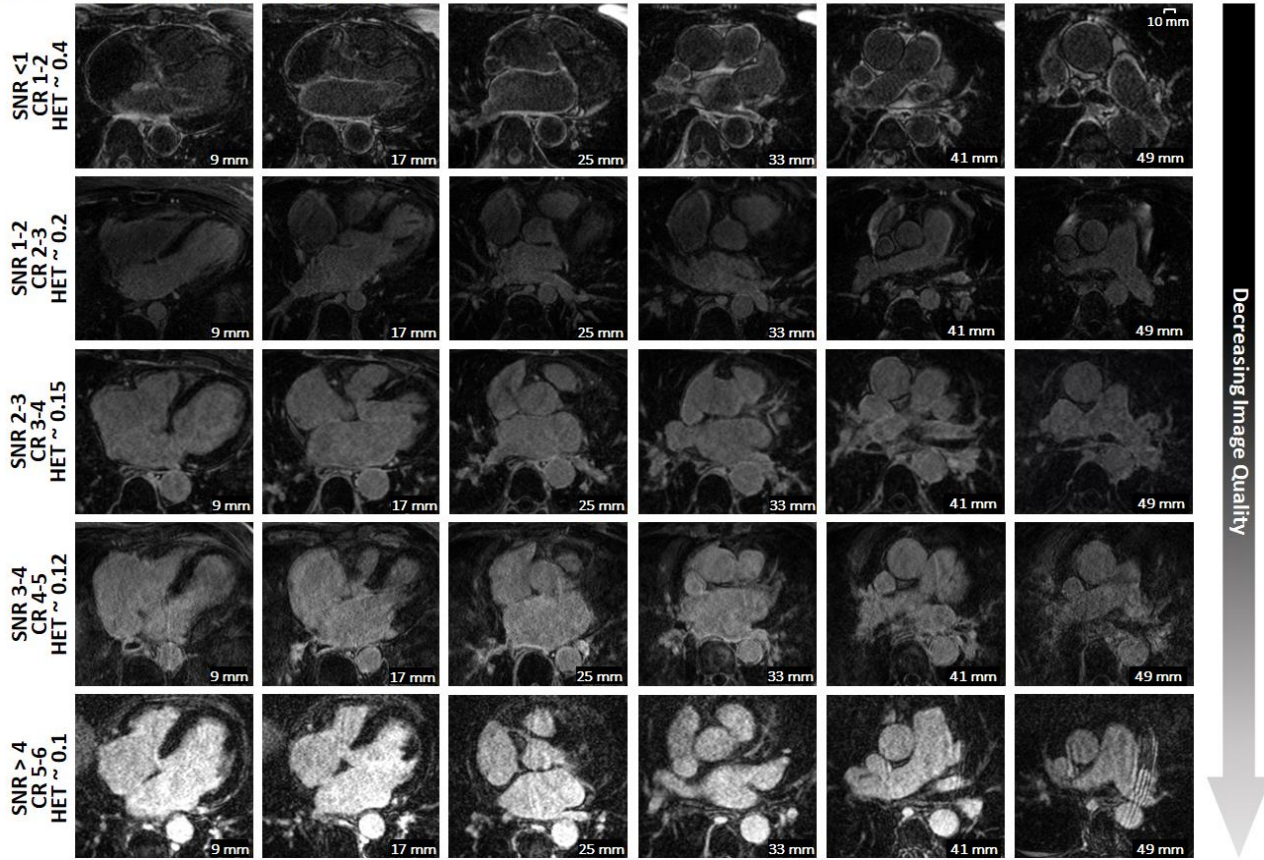
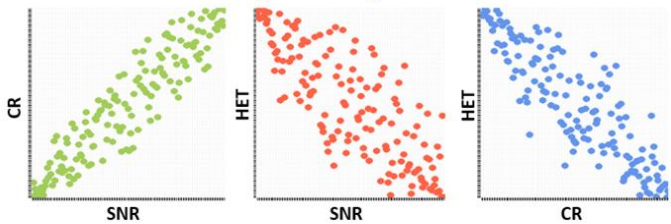
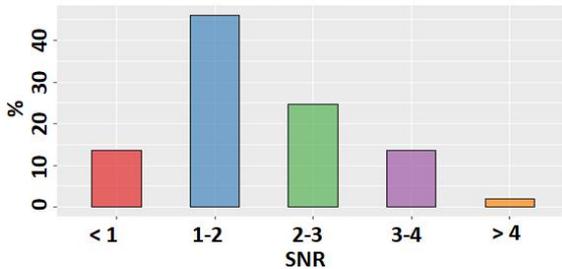
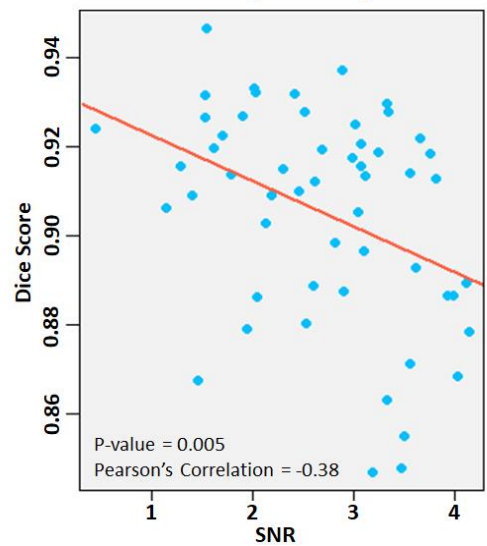

**Figure 2:** Variation in the quality of the late gadolinium-enhanced magnetic imaging (LGE-MRI) dataset used for this study. **A)** Each row represents different LGE-MRIs with a various quality measured using signal to noise ratio (SNR), contrast ratio (CR), and heterogeneity (HET). The data at the top row has the highest quality and the data at the bottom row has the lowest quality. Each column represents 2D LGE-MRI at the same location within a 3D LGE-MRI and is shown with the distance measurements in millimeters. **B)** SNR and CR are positively correlated and are both negatively correlated with HET. **C)** Distribution of the SNR for all the 154 3D LGE-MRI datasets used in the 2018 Left Atrium Segmentation Challenge. **D)** Correlation of the average performance in relation to the LGE-MRI image quality,

demonstrating that lower qualities (higher SNR) result in lower Dice scores.

*2.2. 2018 LA Segmentation Challenge*

The 3D LGE-MRI dataset was randomly split into training (N = 100) and testing (N = 54) sets, with the entire training set published at the start of the challenge for participants to develop their algorithms. The testing set was released without the labels during the evaluation period near the end of the challenge, and participants were ranked based on the accuracy of the testing set predictions submitted to the organizers. The Dice score was used as the only evaluation metric in the challenge for simplicity. However, subsequent analyses with the surface to surface distance and the Euclidean distance error of the LA diameter and volume measurements were conducted after the challenge.

The 17 out of 27 teams who provided their methodologies and performances either in full STACOM papers (Xia et al. 2018, Bian et al. 2018, Vesal, Ravikumar and Maier 2018, Yang et al. 2018, Li et al. 2018, Puybareau et al. 2018, Chen, Bai and Rueckert 2018a, Jia et al. 2018, Liu et al. 2018, Borra et al. 2018, de Vente et al. 2018, Preetha et al. 2018, Qiao et al. 2018, Nuñez-Garcia et al. 2018, Savioli, Montana and Lamata 2018) or online (Huang 2018, Xu 2018) were included in this benchmarking study. Summaries of their methodologies are shown in **Table 1** sorted by the final challenge rankings.

**Table 1**: Summary of methods submitted to the 2018 Left Atrium Segmentation Challenge

| # | Dice (%) | Author | Pre-Processing | 2D/3D | CNNs | Methodology | Post-Processing | Framework |
|---|----------|--------|----------------|-------|------|-------------|-----------------|-----------|
| 1 | **93.2 ± 2.2** | Xia et al.[24] | Down sampling, CLAHE | 3D | 2 | U-Net with additional residual connections to locate ROI, same network to segment ROI | None | PyTorch |
| 2 | **93.1 ± 2.2** | Huang et al.[39] | Down sampling | 3D | 2 | U-Net with additional residual connections, dense connections and dilated convolutions to locate ROI, same network to segment ROI | None | Tensorflow |
| 3 | **92.6 ± 2.2** | Bian et al.[25] | Cropping | 2D | 1 | Dilated ResNet with spatial pyramid pooling to segment images | None | PyTorch |
| 4 | **92.5 ± 2.7** | Vesal et al.[26] | Cropping, CLAHE | 3D | 1 | U-Net with dilated convolutions to segment cropped region | None | Keras |
| 5 | **92.5 ± 2.3** | Yang et al.[27] | Down sampling, cropping | 3D | 2 | Faster-RCNN to locate ROI, U-Net with dense deep supervision to segment the ROI | None | Tensorflow |
| 6 | **92.3 ± 2.9** | Li et al.[28] | None | 3D | 2 | U-Net to locate ROI, U-Net with attention units and hierarchical aggregation units with dilated convolutions to segment the ROI | None | Keras |
| 7 | **92.3 ± 2.3** | Puybareau et al.[29] | Normalization, cropping | 2D | 1 | Fully convolutional network with pre-trained VGG-Net weights and intermediate output maps to segment image | Keep largest component, smoothing | Keras |
| 8 | **92.1 ± 2.6** | Chen et al.[30] | Intensity normalization | 2D | 1 | Multi-task U-Net with an additional classification branch at the center containing spatial pyramid pooling to classify if data is pre or post ablation, training is done on coarse images first then full-sized images to improve feature learning | Dilation and erosion, keep largest component | PyTorch |
| 9 | **91.5 ± 2.6** | Xu et al.[40] | Resize image to multiple scales, cropping | 2D | 1 | Ensemble of different variants of U-Net to segment images of different scales and average the results | None | Tensorflow |
| 10 | **90.7 ± 3.1** | Jia et al.[31] | Normalization, resizing | 3D | 2 | U-Net to locate ROI and generate distance maps, ensemble of U-Nets to segment ROI and distance maps | None | Keras |
| 11 | **90.3 ± 3.2** | Liu et al.[32] | Cropping | 2D | 1 | U-Net to segment cropped region | None | Keras |
| 12 | **89.8 ± 3.4** | Borra et al.[33] | Cropping based on thresholding | 3D | 1 | U-Net to segment cropped region | Keep largest component | Keras |
| 13 | **89.7 ± 3.5** | De Vente et al.[34] | Cropping | 2D | 1 | U-Net to segment patches which were then stitched together to reconstruct the original image | Keep largest component | Keras |
| 14 | **88.7 ± 3.1** | Preetha et al.[35] | Cropping | 2D | 1 | U-Net with deep supervision to segment images | None | Tensorflow |
| 15 | **86.1 ± 3.6** | Qiao et al.[36] | None | - | - | Convert image to probability map, atlas selection, multi-atlas registration, level-set refinement. | None | Non-Deep Learning |
| 16 | **85.9 ± 6.1** | Nuñez-Garcia et al.[37] | None | - | - | Multi-atlas segmentation, shape modeling, clustering to rank similarity of different atria shapes | None | Non-Deep Learning |
| 17 | **85.1 ± 5.1** | Savioli et al.[38] | Cropping, CLAHE, de-noise filters | 3D | 1 | Fully convolutional network to segment entire image volume | None | Torch |

CLAHE, contrast limited adaptive histogram equalization; CNN, convolutional neural network; ROI, region of interest.

*2.3. Algorithm Evaluation*

A range of metrics was used for the benchmarking study. Technical measures for evaluating participants included the Dice score, Intersection over Union (IoU) or Jaccard Index, sensitivity, specificity, Hausdorff distance (HD), and surface to surface distance (STSD). The Dice coefficient is the most commonly used metric for evaluating segmentation accuracy. Given a 3D prediction, *A*, and 3D ground truth, *B*, the Dice score is defined as

$$DICE(A,B) = \frac{2|A \cap B|}{|A| + |B|} \qquad \text{(Eq. 1)}$$

The IoU (or Jaccard index) measures the similarity between a prediction and a ground truth and is defined as

$$IoU(A,B) = \frac{|A \cap B|}{|A \cup B|} \qquad \text{(Eq. 2)}$$

Sensitivity and specificity are used to reflect the success of each algorithm for segmenting the foreground (LA cavity) and the background respectively

$$Sensitivity = \frac{TP}{TP + FN} \qquad \text{(Eq. 3)}$$

$$Specificity = \frac{TN}{TN + FP} \qquad \text{(Eq. 4)}$$

where TP stands for true positive, TN stands for true negative, FP stands for false positive, and FN stands for false negative. The HD measures the local maximum distance between the surfaces of the predicted LA volume and the ground truth. This metric evaluates geometrical characteristics, unlike the Dice or IoU which purely evaluates pixel-by-pixel comparisons. The 3D version of the HD was used in this study to measure the largest error distance of the 3D segmentation defined for a prediction of LA volume, *A*, and ground truth, *B*, as

$$HD(A,B) = \max_{b \in B}\{\min_{a \in A}\{\sqrt{a^2 - b^2}\}\} \qquad \text{(Eq. 5)}$$

where *a* and *b* are all pixels within *A* and *B*. Lastly, the STSD measures the average distance error between the surfaces of the predicted LA volume and the ground truth as

$$STSD(A,B) = \frac{1}{n_A + n_B}\left(\sum_{p=1}^{n_A}\sqrt{p^2 - B^2} + \sum_{p'=1}^{n_B}\sqrt{p'^2 - A^2}\right) \qquad \text{(Eq. 6)}$$

where $n_A$ is the number of pixels in *A*, $n_B$ is the number of pixels in *B*, and *p* and *p'* describe all points in *A* and *B*.

Biological measures for evaluating performance included the error of the LA anterior-posterior diameter and the error of the 3D LA volume between predictions and ground truth since LA diameter and volume are the two widely used clinical measures during the clinical diagnosis and treatment of patients with AF. The LA diameter, measured in millimeters, is calculated by finding the maximum Euclidean distance along the *x*-axis of each MRI scan to estimate the distance from the anterior LA to the posterior. The LA volume, measured in $cm^3$, is calculated by summing the total number of positive (LA cavity) pixels. Participant evaluation and ranking using these metrics are summarized in **Figure 4D-G**.

Throughout the study, statistical significance was assessed using the two-tailed *T*-test to compare the performances of participant sub-groups as well as individual algorithms during analyses.

## 3. Results

*3.1. Performance of Submitted Algorithms*

We first examined the overall performance of the different methodology categories from the 17 submitted algorithms. CNNs were the most popular choice as it was used by 15 teams, and on average, substantially outperformed the other two teams which used traditional atlas-based segmentation methods (92.5% vs 85.1% Dice score, $p < 0.05$, **Figure 3A**). Participants used a variety of deep learning libraries for implementing their CNN pipelines including Keras (N = 7), Tensorflow (N = 4), PyTorch (N = 3), and Torch (N = 1). It was observed that the teams implemented with PyTorch performed better (**Figure 3B**). Out of the 15 teams using CNNs, 12 teams proposed CNNs designs based on the popular U-Net architecture (Ronneberger et al. 2015) whilst the other three 3 implemented non-U-Net designs. We observed that the teams using U-Net based CNNs had superior performances (92.4% vs 89.3% Dice score, $p < 0.05$) including Xia et al. (Xia et al. 2018) and Huang et al. (Huang 2018) who were ranked $1^{st}$ and $2^{nd}$ in the challenge (**Figure 3C**). The majority of teams using the U-Net architecture implemented further enhancements to the original architecture in an attempt to improve the segmentation performance. This involved the use of additional residual connections (Xia et al. 2018), replacing all layers with dilated convolutions (Vesal et al. 2018), improved methods of training such as the use of custom loss functions, deep supervision (Yang et al. 2018), multi-task learning (Chen et al. 2018a), and attention mechanisms throughout the network (Li et al. 2018). The three teams which did not use U-Net as a baseline approach implemented enhanced versions of existing architectures such as ResNet (Bian et al. 2018, Szegedy et al. 2017), VGGNet (Simonyan and Zisserman 2014, Puybareau et al. 2018), and Fully-CNNs (Long et al. 2015, Savioli et al. 2018, Xiong et al. 2019) which have been widely used on the ImageNet database (Deng et al. 2009).

We then evaluated 2D versus 3D approaches based on the 15 submitted CNN algorithms. Since the challenge data was 3D, 8 out of the 15 teams proposed 3D CNNs which simply performed direct 3D segmentation on each set of 3D LGE-MRIs. On the other hand, 2D CNNs used by seven teams segmented each image slice of the image volume along the **z**-axis and stacked the individual segmentations together to obtain the final 3D results. Summary statistics show that there is no significant difference between the 2D and 3D CNN approaches when the CNN architecture and setup are not considered (92.1% vs 92.5% Dice score, $p = 0.82$, **Figure 3D**), even though the 3D CNNs were used by 4 of the top 5 teams. We also observed five teams utilizing a double, sequentially used CNNs in their pipeline to improve the segmentation performance compared to methods that only contain a single CNN (N = 10). Through this enhancement, the former approach achieved a significantly better average Dice score of ~92.8% compared to single CNN methods which obtained a Dice score of ~90.3% on average (**Figure 3E**).

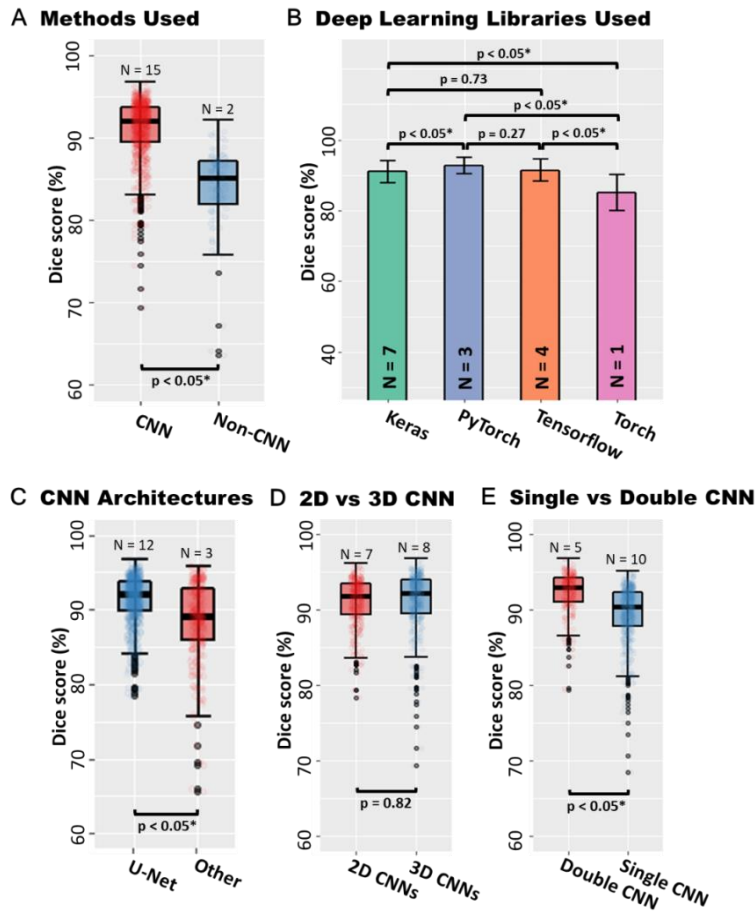

**Figure 3:** Comparative summaries of the 17 algorithms included in this benchmarking study, a well representative subset of the 27 teams that participated in the 2018 Atrial Segmentation Challenge. **A)** The 15 methods utilizing convolutional neural networks (CNNs) outperformed the two traditional multi-atlas based methods with statistical significance. **B)** A range of deep learning libraries were used by the 15 teams to implement CNNs. **C)** The 12 teams using U-Net architectures in their CNN pipelines outperformed the 3 teams using ResNet, VGGNet, and Fully-CNNs with statistical significance. **D)** Approximately an equal number of teams used methods consisting of 2D CNNs compared to methods consisting of 3D CNNs. However, there was no statistically significant difference in segmentation accuracy between the two groups. **E)** Participants using a double, sequentially used CNNs achieved significantly higher accuracy than those using a single CNN for segmentation.

Of the methods discussed in our study, the double, sequentially used CNNs, termed as *double CNN* throughout this study, involved the first CNN automatically detecting the region of interest (ROI) from LGE-MRIs and the second CNN performing regional segmentation of the LA from the ROIs (**Figure 4A**). Alternatively, the *single CNN* method focused solely on using one CNN for direct segmentation of the LA from either the original dataset or ROIs cropped at a consistent location across all input images (**Figure 4B-C**). To further examine the 15 proposed CNN methodologies, we have regrouped the approaches into three categories: double 3D CNNs (N = 5), single 2D CNNs (N = 7), and single 3D CNNs (N = 3). It is noted that no team proposed a double 2D CNN method. The average accuracy of the three categories of CNN methods were evaluated with both technical and biological performance metrics: Double 3D CNN methods significantly outperformed single 2D and 3D CNN methods in terms of the Dice score (92.8% vs 91.1% and 89.9%, $p < 0.05$), the surface to surface distance (0.75mm vs 0.85mm and 1.1mm, $p < 0.05$), LA diameter error (2.7% vs 3.2% and 4.3%, $p < 0.05$), and LA volume error (4.5% vs 4.9% and 6.2%, $p < 0.05$) (**Figure**

**4D-G**). The success of the double CNN methodology can also be seen in the challenge rankings as this workflow was utilized by the top 2 teams as well as 4 of the top 6 teams which achieved over 92% Dice score.

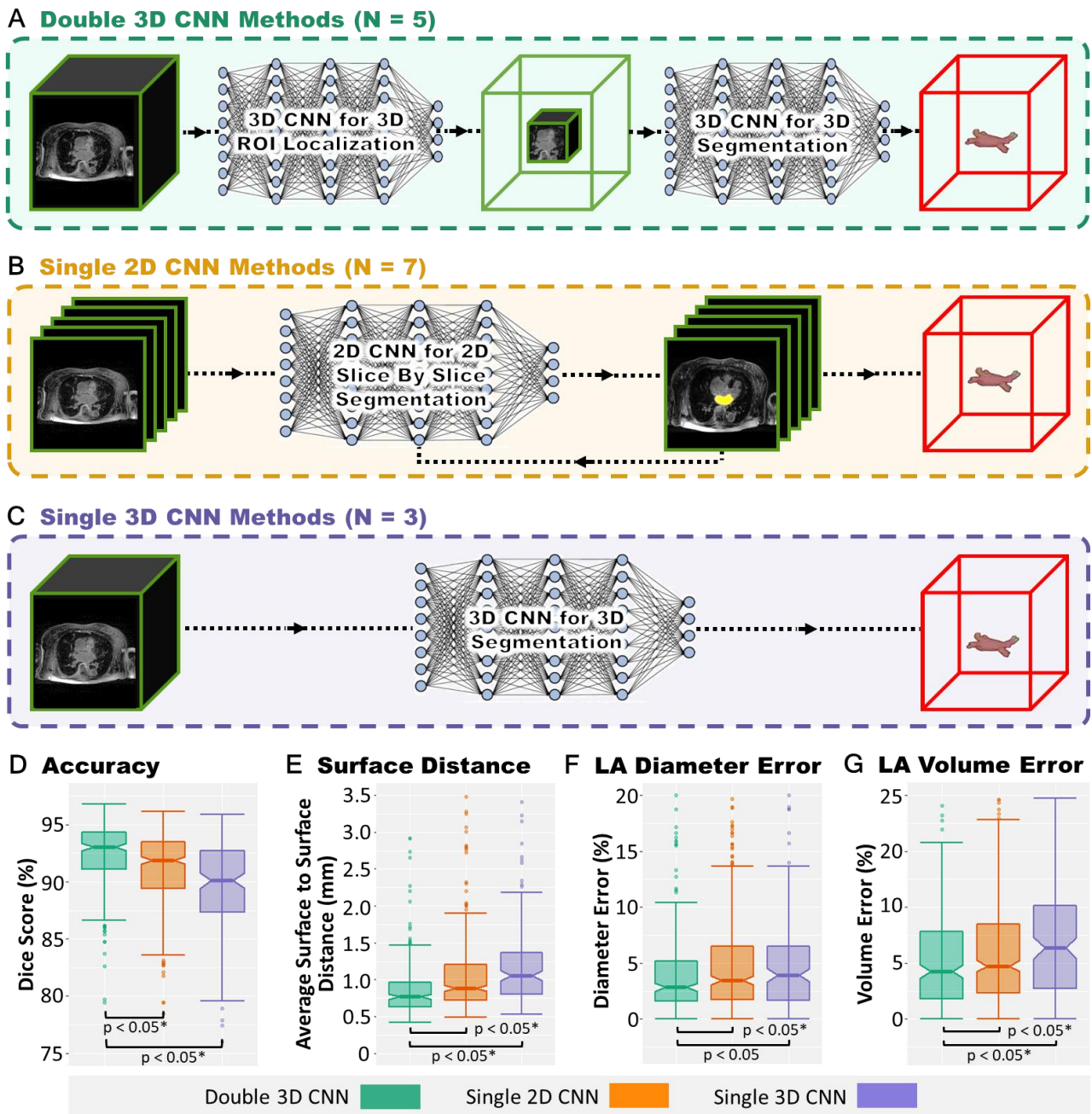

**Figure 4:** Detailed evaluations of the performance of the 15 convolutional neural network (CNN) pipelines submitted to the challenge grouped into three general categories depending on how the CNNs were applied to segment the left atrium (LA) from the late gadolinium-enhanced magnetic resonance imaging (LGE-MRI). **A)** Double 3D CNN methods consisted of one CNN to detect a 3D region of interest (ROI) from the LGE-MRIs and a second CNN to segment LA from the 3D ROI. **B)** Single 2D CNN methods consisted of one 2D CNN which performed slice-by-slice segmentation of each LGE-MRI. The 3D LA was then reconstructed by stacking slice-by-slice segmentation. **C)** Single 3D CNN methods consisted of a single 3D CNN to segment the LA from the LGE-MRI volume directly. Dice score (**D**), the average surface to surface distance (**E**), LA diameter error (**F**) and LA volume error comparisons (**G**) showed the superiority of 3D double CNN methods among the three different categories of CNN workflows.

*3.2. Top Performing CNN Methodologies*

Methods involving double 3D CNNs were shown to have the best overall performance. Interestingly, all five of the double 3D CNN methods adopted U-Net as the baseline architecture, though significant improvements were added to the baseline approach by these participants. These included the addition of residual connections into the U-Net by Xia et al. (Xia et al. 2018) (**Figure S1**), Huang et al. (Huang 2018) (**Figure S2**), and Li et al. (Li et al. 2018) (**Figure S3**). Specifically, residual connections were added to each block of two to three sequential convolutional layers along the entire length of the networks to improve gradient flow during backpropagation when training the CNNs. The type of residual connections varied from a simple connection without any additional operations to more advanced connections containing convolutional and pooling layers. Dense connections were also seen in the method proposed by Huang et al. (Huang 2018) along with dilated convolutions to improve the receptive field of the CNN. On the other hand, Yang et al. (Yang et al. 2018) did not alter the U-Net architecture but instead elected for an improved dense supervision training scheme and a customized loss function (**Figure S4**). The proposed loss function was an ensemble of the Dice score, pixel thresholding to improve sensitivity, and an overlap metric for improving segmentations at boundary locations. Apart from the double 3D CNN methods, Bian et al. (Bian et al. 2018) and Vesal et al. (Vesal et al. 2018) also performed highly with single 3D CNN methods. The effectiveness of their CNNs could be potentially attributed to the use of dilated convolutions, allowing them to outperform all other single CNN methods.

To gain further insights into the top performing CNN pipelines, control experiments were performed in this benchmarking study on the winning approach by Xia et al. (Xia et al. 2018) ($p < 0.05$ when comparing all technical metrics against other teams) to examine the factors contributing to their superior performance (**Figure 5**). It is noted that since most top-ranking algorithms utilized similar U-Net based designs, experimental observations derived from the approach by Xia et al. (Xia et al. 2018) would also apply to other methods with similar algorithm setups. For the purposes of analysis, the top algorithm was selected based on its simplicity and light-weight nature. A summary of the participants' double 3D CNN pipeline is shown in **Figure 5A** and additional experiments were conducted to evaluate the efficacy of the second CNN network for performing segmentation. Hyper-parameter tuning experiments in **Figure 5B** revealed that the extra residual connections added to the U-Net architecture increased the Dice score by 0.7%. The increased receptive field using $5 \times 5 \times 5$ convolutional kernels significantly improved the Dice score by 4% compared to the widely used $3 \times 3 \times 3$ kernels. The use of the Dice loss improved the accuracy by 2.1% over the traditional cross-entropy loss which does not account for the major class imbalance present in the dataset. Although not statistically significant, dropout and parametric rectified linear unit (PReLU) further improved performance by approximately 0.5%. Color-intensity normalization or contrast limited adaptive histogram equalization (CLAHE), used by several teams improved the Dice score by 0.7% (**Figure 5C**). Expectedly, standard data augmentation techniques such as random rotation, elastic deformations, perspective scaling, and random flipping improved the performance by over 2% (**Figure 5D**), while other schemes such as blurring, affine transformations and sheering did not result in any significant improvements. We also observed that online data augmentation was potentially more effective than offline (**Figure 5E**), likely due to the larger variety of data generated on-the-fly during training.

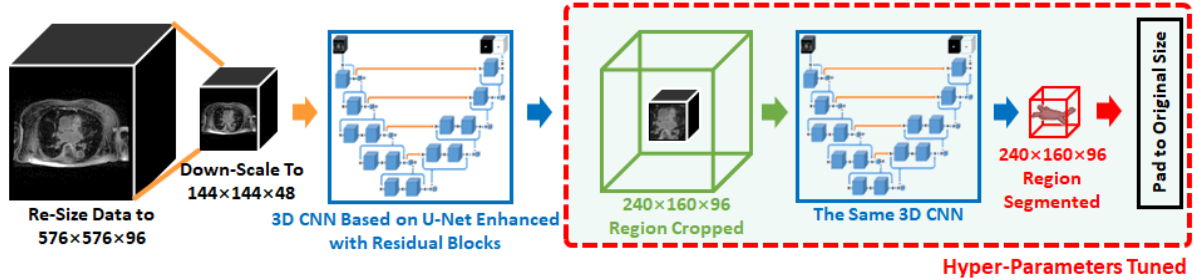
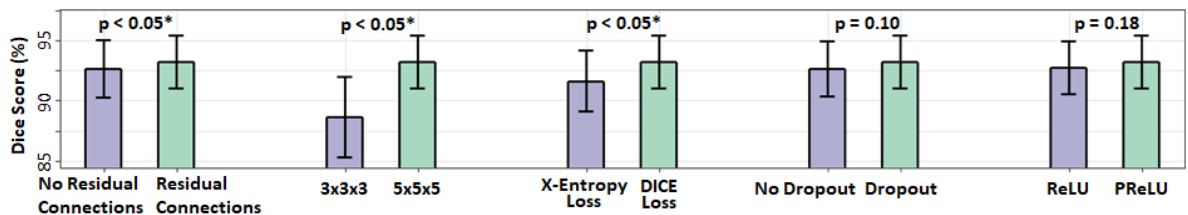
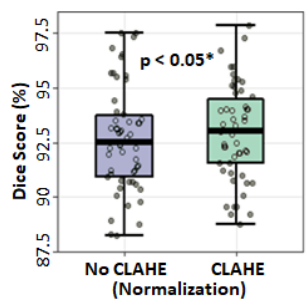
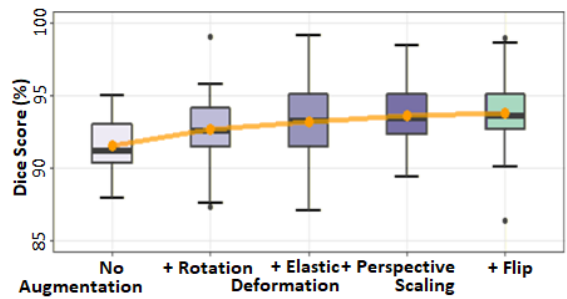
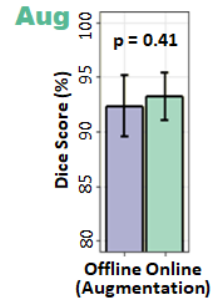

**Figure 5:** Post-challenge analysis of the winning method by Xia et al. (Xia et al. 2018) in the 2018 Left Atrium (LA) Segmentation Challenge demonstrates the optimality of their approach. **A)** Schematic summary of the 3D double convolutional neural network (CNN) approach. Both CNNs consisted of a U-Net architecture enhanced with batch-normalization in each layer and residual connections along the length of the network. The first CNN detected the centroid of the ROI from a down-sampled version of the initial late gadolinium-enhanced magnetic resonance imaging (LGE-MRI). A 240×160×96 region centered in the LA cavity was the output and was then processed by the second CNN to segment LA in 3D. The output was padded to obtain the original resolution of the input LGE-MRI. **B)** Hyper-parameter tuning of the U-Net architecture in the second CNN (red box) showed that all parameters used by the winning team were optimal through post-challenge analysis. **C)** The effect of contrast limited adaptive histogram equalization (CLAHE) for normalizing each LGE-MRI during pre-processing on the performance of the CNN. **D)** Data augmentation led to an incremental improvement as more augmentation methods were increasingly added, showing an increasing trend in accuracy. **E)** Comparisons of offline and online data augmentation schemes showing that online augmentation resulted in marginally higher performances.

### 3.3. Key Factors Influencing the Performance of Double CNN Approaches

We performed further analyses to target the key factors contributing to the success of double CNN methods compared to single CNNs. Since the optimality of the second CNN was shown through experimentation in the previous subsection, we focused on analyzing the key factors of the first CNN for influencing the final segmentation accuracy (**Figure 6**). Firstly, it was observed that the ROI yielded as a result of the first CNN in a double CNN method was consistently centered on the LA (**Figure 6A**). Without the ROI detection procedure undergone by the first CNN, the center of the LA would potentially have a ~100 pixel or ~17%, shift in position from that of the original input images. Our experiments show a decreasing trend in overall CNN performance as the LA (ROI) was purposely shifted away from the center of the image patch as an input for the second CNN, suggesting that centering the LA is extremely important to obtain superior accuracies (**Figure 6B** and **C**). Secondly, we observed that the smaller the image patch of ROI as the output of the first CNN, the higher final segmentation accuracy. This relationship was true for all ROI sizes that were greater than 240×160 (dashed cyan line in **Figure 6D**). The decreased input size of ROIs generated from the first CNN of the double CNN methods reduced the class imbalance as there were significantly fewer background pixels present in the original input images, resulting in better performances as seen in our experiments using input sizes with X/Y dimensions of 240×160 to 400×400. However, our experimentation also showed that the Dice accuracy of the CNN decreased when the size of the ROI was less than 240×160 even though the LA was fully contained within the ROI. We postulate that the observed decrease in performance could be attributed to the boundary of the LA being too close to the edge of the ROI inputted into the CNN. Furthermore, U-Net is known to perform poorly when segmenting boundary regions (Ronneberger et al. 2015).

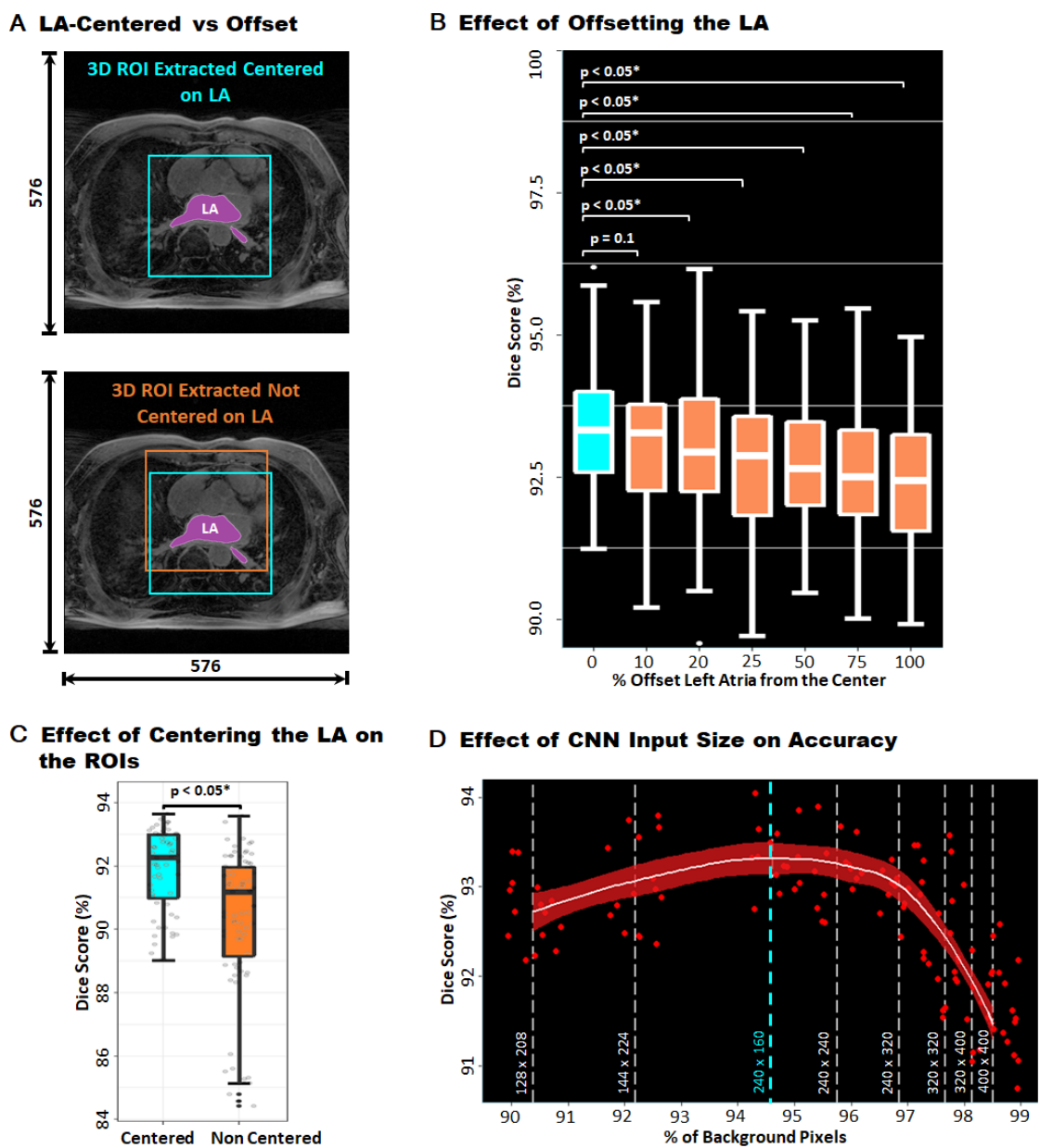

**Figure 6:** Analysis of the key factors of the first convolutional neural network (CNN) in the double CNN methods in enhancing left

atrium (LA) segmentation performance. **A)** Illustration of the region of interest (ROI) extracted from the double CNN methods centered on the LA (Top) and ROI extracted not centered on the LA (Bottom). The ROI not centered on the LA can be offset from the center of the LA by as much as 100 pixels in any direction (orange box). **B)** Displacing the LA by a distance in the ROI from the center of the patch reduced segmentation accuracy. A 100% offset implies the LA is pressed against the side of the patch without any loss of LA pixels. **C)** Performance summary comparing the methods proposed in the challenge which used double CNN pipelines to achieve an LA centered patch and the methods which had non-centered patches due to single CNN pipelines. **D)** To achieve the best accuracy, the size of the ROI should be sufficiently small. Effect of the input patch size of a CNN on the segmentation accuracy showing the peak accuracy at a patch size of 240×160 as proposed by the winners of the challenge. The percentage of background pixels is computed along the tested patch sizes to convey the degree of class imbalance in the CNN inputs where a higher percentage represents a more severe class imbalance.

## 4. Discussion and Conclusions

To our knowledge, this is the largest study of this kind to systematically benchmark various approaches submitted to a global challenge for LA segmentation from 3D LGE-MRIs. These approaches were developed based on the largest and highest quality 3D cardiac LGE-MRIs dataset and labels in the world provided by our 2018 LA segmentation challenge. The benchmarking study includes the approaches developed by 17 groups, a well representative subset of the 27 teams participated. While prior efforts exist (Tobon-Gomez et al. 2015, Bernard et al. 2018, Karim et al. 2018), our benchmarking study makes several significant contributions. Firstly, our study is one of few that investigate methods of LA segmentation directly from LGE-MRIs, which is more challenging than segmenting non-contrast images due to the attenuated color contrast causing a lack of clarity between the atrial tissue and background pixels. Furthermore, we attracted more participants than prior studies and our dataset is multi-fold larger in comparison to previous challenges. In addition, our task of LA segmentation is significantly more difficult than prior established challenges which focused on other larger cardiac structures, given the relatively small LA chamber and thin wall (Peng et al. 2016). As a result, the methods proposed are more superior and robust as seen from the diversity of the proposed approaches and the various performance metric evaluations. Lastly, our study delves deeper into the key factors of CNN approaches that optimize segmentation performance through extensive post-challenge analyses on the 17 submitted approaches.

*4.1. Characteristics of Top-Performing CNNs for Segmentation*

The key findings of this study are multifold. First of all, our study found that CNN approaches, notably those based on the U-Net architecture achieved better performance compared to traditional atlas-based methods and other CNN architectures, similar to previous studies (Bernard et al. 2018). Furthermore, by observing the methods proposed by the top five teams, we believe that additional residual connections and improved methods of optimization, such as custom loss functions, resulted in higher performance. Our post-challenge analysis also showed that data augmentation methods and color-intensity or contrast normalization enhanced feature learning improved accuracy. More interestingly, 2D and 3D CNN methods had comparable accuracies, showing that analyzing each LGE-MRI slice in 2D and ignoring the 3D continuity in the data achieves similar results while requiring significantly less computation and memory cost due to the smaller number of parameters. In particular, we discovered that a double, sequentially used CNN architecture achieved far superior segmentation results than a single CNN approach for both technical and biological performance metrics. The effectiveness of the double CNN methods relies on two key elements. Firstly, the first CNN automatically detects the location of the ROI resulting in the image patch extracted to be centered on the ROI and sufficiently small to include minimal background pixels. Secondly, the second CNN effectively performs detailed regional segmentation of the ROI from the extracted image patch. This pipeline can be trained end-to-end and is far more computationally efficient compared to the state-of-the-art ensemble methods that have won previous segmentation challenges (Bernard et al. 2018). We believe that the superior framework of the double CNN method can be applied to any problem with severe class imbalance such as the segmentation of small structures from larger input images. The fully data-driven nature of the double CNN approaches also means that they are widely adaptable to other image datasets such as 3D contrast-enhanced CTs and non-contrast medical images.

*4.2. Comparison of Evaluation Metrics*

While **Figure 4** showed that 3D double CNNs achieved the highest performance on average regardless of the evaluation metric used, comparisons of individual algorithms showed a slight discrepancy in the overall rankings (**Figure 7**). Results showed that although the most widely used segmentation metrics consisting of the Dice score, Jaccard Index/IoU, and the surface to surface distance produced fairly consistent rankings amongst the top 5 contestants, the Hausdorff distance, sensitivity, and specificity produced significantly different rankings in comparison. A potential explanation of this discrepancy is that the Hausdorff distance measured the most extreme errors, while sensitivity and specificity do not consider both positive and negative pixels simultaneously, leading to biased measurements. This also likely explains why these metrics are not commonly used for evaluating segmentation accuracies. While the conclusions drawn from the study based on the experimental results were validated with multiple technical and biological metrics, these summary results still suggest the ongoing need for an improved global definition of the exact meaning of "high segmentation accuracy".

| Author | Dice (%) | IoU (%) | Sensitivity (%) | Specificity (%) | Hausdorff Distance (mm) | Surface to Surface Distance (mm) | P-Value |
|---|---|---|---|---|---|---|---|
| Xia et al. | 93.2 (2.2) | 87.4 (3.8) | 93.6 (3.4) | 99.952 (0.032) | 8.892 (4.160) | 0.748 (0.224) | 0.050* |
| Huang et al. | 93.1 (2.2) | 87.2 (3.8) | 93.7 (3.5) | 99.949 (0.034) | 8.495 (4.088) | 0.754 (0.224) | 0.043* |
| Bian et al. | 92.6 (2.2) | 86.9 (3.7) | 93.3 (3.5) | 99.951 (0.034) | 9.213 (5.319) | 0.759 (0.226) | 0.012* |
| Yang et al. | 92.5 (2.7) | 86.1 (4.4) | 94.3 (3.3) | 99.932 (0.042) | 9.759 (5.981) | 0.850 (0.332) | 0.099 |
| Vesal et al. | 92.5 (2.3) | 86.0 (3.9) | 91.9 (4.5) | 99.954 (0.029) | 9.444 (4.671) | 0.817 (0.240) | 0.101 |
| Lee et al. | 92.3 (2.9) | 85.9 (4.9) | 92.2 (5.2) | 99.950 (0.030) | 10.593 (7.109) | 0.897 (0.442) | 0.230 |
| Puybareau et al. | 92.3 (2.3) | 85.7 (3.9) | 91.7 (4.1) | 99.952 (0.030) | 9.812 (5.676) | 0.854 (0.252) | 0.251 |
| Chen et al. | 92.1 (2.6) | 85.4 (4.3) | 91.8 (4.3) | 99.949 (0.034) | 8.603 (4.710) | 0.854 (0.245) | 0.148 |
| Xu et al. | 91.5 (2.6) | 84.5 (4.3) | 90.0 (3.9) | 99.955 (0.025) | 10.912 (5.000) | 0.911 (0.258) | 0.010* |
| Jia et al. | 90.7 (3.1) | 83.2 (5.1) | 91.2 (5.4) | 99.937 (0.038) | 10.683 (5.976) | 1.087 (0.496) | 0.029* |
| Liu et al. | 90.3 (3.2) | 82.5 (5.2) | 89.2 (4.8) | 99.945 (0.037) | 8.694 (3.841) | 1.304 (0.845) | 0.028* |
| Borra et al. | 89.8 (3.4) | 81.7 (5.4) | 91.0 (6.9) | 99.992 (0.034) | 12.038 (5.586) | 1.155 (0.454) | 0.005* |
| De Vente et al. | 89.7 (3.5) | 81.5 (5.6) | 88.9 (6.1) | 99.938 (0.032) | 9.766 (4.905) | 1.132 (0.349) | 0.001* |
| Preetha et al. | 88.7 (3.1) | 79.9 (4.9) | 94.9 (3.0) | 99.873 (0.048) | 8.570 (3.486) | 2.608 (1.933) | 0.0001* |
| Qiao et al. | 86.1 (3.6) | 75.8 (5.4) | 84.7 (6.0) | 99.923 (0.041) | 11.834 (4.510) | 1.449 (0.404) | 0.030* |
| Nuñez-Garcia et al. | 85.9 (6.1) | 75.8 (8.7) | 84.5 (9.1) | 99.250 (0.039) | 12.690 (5.159) | 1.473 (0.599) | 0.035* |
| Savioli et al. | 85.1 (5.1) | 74.4 (7.4) | 83.6 (8.7) | 99.917 (0.048) | 14.659 (6.789) | 1.611 (0.645) | 0.021* |

Lower Performance ← → Higher Performance

**Figure 7:** Summary of the rankings for all participants under different technical metrics. Metrics included the Dice score, Intersection over Union (IoU/Jaccard Index), sensitivity, specificity, Hausdorff distance (HD) and average surface to surface distance performance measures. The color intensities reflect the rankings of the teams such that darker colors represent teams with higher performances. P-values for the statistical significance of each team when comparing all technical metrics to other teams are shown in the last column, with statistically significant values marked with asterisks (*).

*4.3. Error Analysis and Future Work*

Error analysis of the top methods showed that although the 3D visualizations of the predicted LAs were similar to the ground truths, there are still potential improvements to be made. Slice-by-slice analysis of the 3D predictions from the top team showed that the method performed poorly when segmenting the regions containing the PVs located at the superior slices of the 3D LGE-MRIs and the mitral valve at the bottom connecting the LA with the left ventricle (**Figure 8**). The errors at the mitral valve were attributed to the fact that there are no clear landmarks to separate the two chambers. This leads the experts to label this region with a flat plane which potentially contains large inter-observer variability, making it difficult to be reproduced by the CNNs. On the other hand, the errors at the PVs could be explained by the fact that these structures are often very small in size and vary greatly in shape between patients, making them difficult to detect. The inherently poor LGE-MRI qualities were also a factor which impacted the segmentation performance due to the low contrast of the images. Analyses showed that the average accuracy for each LGE-MRI was directly correlated to the quality of the particular LGE-MRI measured in the signal-to-noise ratio for all approaches (**Figure 2D**).

In the future, these issues could potentially be mitigated with an increased number of LGE-MRIs and the use of multi-center LGE-MRIs to improve the robustness of the CNNs on a greater diversity of datasets. Further, we would also like to extend current methodologies in this study to a concurrent multi-label problem, such as the segmentation of both atrial chambers and cavities simultaneously. The concept of 2D double CNNs would also be another interesting direction for future research as it was not proposed by any of the teams, and may potentially be utilized to further improve segmentation accuracies on more difficult tasks such as bi-atrial chamber segmentation.

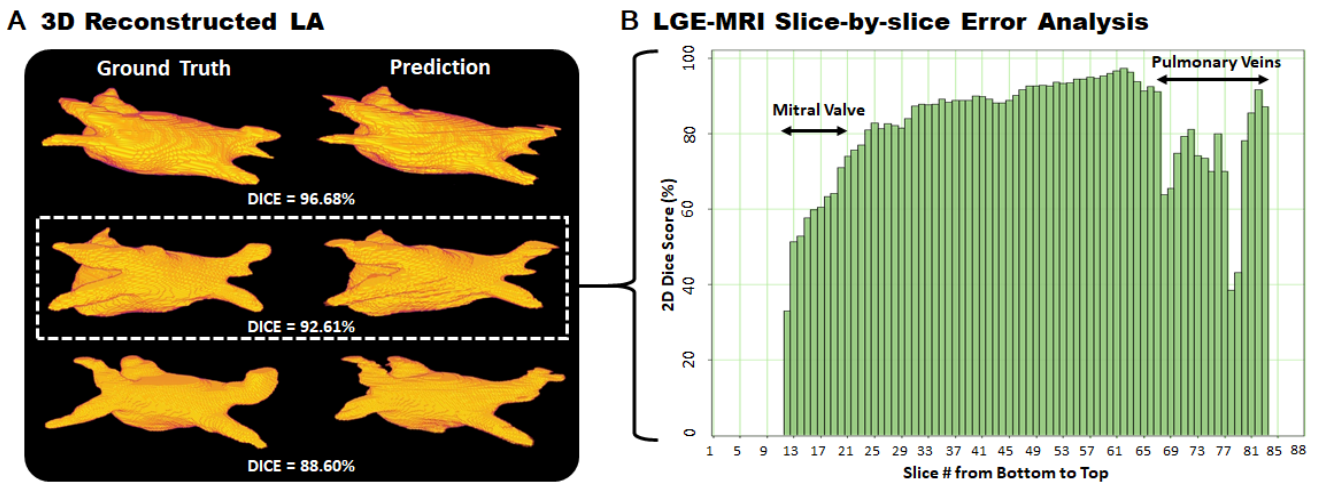

**Figure 8:** Error analysis of predictions generated from the winning method by Xia et al.(Xia et al. 2018) in the 2018 Left Atrium (LA) Segmentation Challenge. **a)** Visualization of predicted 3D LA reconstructions with a high, medium, and low Dice scores. **b)** Slice-by-slice evaluation of the predicted LA reconstruction with a 3D Dice score of 92.61% in segmentation accuracy across each of the 88 slices in a 3D late gadolinium-enhanced magnetic resonance imaging (LGE-MRI). The most erroneous slices were at the bottom and top of the LA corresponding to the mitral valve and pulmonary veins, respectively. Slices 1 to 11 and 83 to 88 did not have any LA pixels; hence, did not have any accuracy values. Note that the 2D Dice score used does not average out to the overall 3D Dice score due to the different dimensionality of the accuracy assessment.

*4.4. Conclusions*

Segmentation of cardiac images, particularly LGE-MRIs widely used for visualizing diseased cardiac structures, is a crucial first step for clinical prognosis and treatment. This study describes the 2018 LA segmentation challenge which provides 154 3D LGE-MRIs, and elaborates on the subsequent analysis performed on the algorithms submitted for LA segmentation. Our study found U-Net CNNs achieved the best performance on average, and the use of additional residual connections and advanced methods of optimization such as custom loss functions resulted in higher performance. Analysis also showed that while 2D and 3D CNN methods had comparable accuracies, double, sequentially used CNNs achieved far superior segmentation results compared to single CNNs regardless of the input dimension. Detailed practical considerations when designing CNNs pipelines were also investigated through our extensive hyper-parameter tuning experiments, gaining insights into the process of obtaining state-of-the-art accuracies. We believe that findings from this study can potentially be extended to other imaging datasets and modalities, having an impact on the wider medical imaging community. This large-scale benchmarking study makes a significant step towards much-improved segmentation methods for cardiac LGE-MRIs, and will serve as an important benchmark for evaluating and comparing the future works in the field.

**Acknowledgements**

The authors would like to thank Nvidia, MedTech CoRE New Zealand, and Arterys for providing prizes for the winners of the 2018 LA Segmentation Challenge. Z.X. and J.Z. are grateful for Nvidia for donating Titan-X Pascal GPU for algorithm development and testing, and The NIH/NIGMS Center for Integrative Biomedical Computing (CIBC) at the University of Utah for providing the LGE-MRI dataset. This work was funded by the Health Research Council of New Zealand [grant number 16/385].

**Author Contributions**

Z.X and J.Z organized the challenge, analyzed the results, and produced the figures and text for the study. All remaining authors (Q.X, Z.H, N.H, S.V, N.R, A.M, C.L, Q.T, W.S, E.P, Y.K, T.G, C.C, W.B, D.R, L.X, X.Z, X.L, S.J, M.S, D.B, A.M, C.C, R.K, C.V, M.V, C.J.P, S.E, M.Q, Y.W, Q.T, M.N.G, O.C, N.S, P.L, Y.L, K.W) were challenge participants and provided descriptions of their methodologies and results and revised the final manuscript.

**Conflict of Interest**

The authors declare no conflict of interests.